%% file: main.tex
\documentclass[conference]{IEEEtran}
\IEEEoverridecommandlockouts

\usepackage{cite}
\usepackage{amsmath,amssymb,amsfonts}
\usepackage{algorithmic}
\usepackage{graphicx}
\usepackage{textcomp}
\usepackage{xcolor}
\usepackage{hyperref}
\usepackage{booktabs} 
\usepackage{enumitem}
\usepackage{graphicx}
\usepackage{multirow} 

\usepackage{bm}
\def\BibTeX{{\rm B\kern-.05em{\sc i\kern-.025em b}\kern-.08em
    T\kern-.1667em\lower.7ex\hbox{E}\kern-.125emX}}

\newcommand{\mypar}[1]{\noindent{\bf #1}}

\begin{document}

\title{EMO-Reasoning: Benchmarking Emotional Reasoning Capabilities in Spoken Dialogue Systems}

\author{%
  \IEEEauthorblockN{%
    \parbox{\linewidth}{\centering
      Jingwen Liu\IEEEauthorrefmark{4}$^*$, %
      Kan Jen Cheng\IEEEauthorrefmark{2}$^*$, %
      Jiachen Lian\IEEEauthorrefmark{2}\thanks{First two authors contributed equally. Project lead / Corresponding author: \texttt{jiachenlian@berkeley.edu}}, %
      Akshay Anand\IEEEauthorrefmark{2}, Rishi Jain\IEEEauthorrefmark{2}, Faith Qiao\IEEEauthorrefmark{2}, Robin Netzorg\IEEEauthorrefmark{2},\\
      Huang-Cheng Chou\IEEEauthorrefmark{5}, Tingle Li\IEEEauthorrefmark{2}, Guan-Ting Lin\IEEEauthorrefmark{3}, Gopala Anumanchipalli\IEEEauthorrefmark{2}%
    }
  }
  \vspace{2mm}
  \IEEEauthorblockA{\normalsize  
    \IEEEauthorrefmark{4}Zhejiang University \quad
    \IEEEauthorrefmark{2}UC Berkeley \quad
    \IEEEauthorrefmark{3}National Taiwan University
    \IEEEauthorrefmark{5}University of Southern California \\[0.1em]
    \small{\tt \url{https://berkeley-speech-group.github.io/emo-reasoning/}}
  }
}

\maketitle

\input{tex/abstract}

\begin{IEEEkeywords}
Emotion Reasoning, Benchmark, Spoken Dialogue Systems, Emotion Perception
\end{IEEEkeywords}

\input{tex/intro}
\input{tex/method}

\input{tex/experiment}
\input{tex/conclusion}

\bibliographystyle{IEEEtran}
\bibliography{main}
\end{document}

%% file: tex/abstract.tex
\begin{abstract}

Speech emotions play a crucial role in human-computer interaction, shaping engagement and context-aware communication. Despite recent advances in spoken dialogue systems, a holistic system for evaluating emotional reasoning is still lacking. To address this, we introduce EMO-Reasoning, a benchmark for assessing emotional coherence in dialogue systems. It leverages a curated dataset generated via text-to-speech to simulate diverse emotional states, overcoming the scarcity of emotional speech data. We further propose the Cross-turn Emotion Reasoning Score to assess the emotion transitions in multi-turn dialogues. Evaluating seven dialogue systems through continuous, categorical, and perceptual metrics, we show that our framework effectively detects emotional inconsistencies, providing insights for improving current dialogue systems. By releasing a systematic evaluation benchmark, we aim to advance emotion-aware spoken dialogue modeling toward more natural and adaptive interactions.

\end{abstract}

%% file: tex/intro.tex
\section{Introduction}

Spoken communication relies on more than just the words we use. The tone of voice, emotional expression, and timing all play key roles in conveying and interpreting meaning \cite{quinto2013emotional, mullennix2002effects}. While text-based language models have advanced significantly, their ability to handle the nuances of spoken emotion remains limited. In real-world conversations, a speaker’s emotional state can shift over time, influencing both how they express themselves and how others respond. Recognizing and managing these shifts is a critical step toward more intuitive and empathetic human-computer interactions.

Although many spoken dialogue models (SDMs) excel at understanding linguistic content, they often overlook the rational use of emotion. For example, showing empathy when a user is distressed or sharing joy to build rapport (Figure \ref{fig:metrics}). Prior research has introduced benchmarks for tasks such as spoken question answering \cite{hassid2024textually, nachmani2023spoken, shih2023gsqa, lin2024align, lin2022dual}, instruction following \cite{huang2024dynamic, huang2024dynamic2}, style-based conversation \cite{voxdialogue, ao2024sd, styletalk, lin2024can, lin2024paralinguistics}, duplex systems~\cite{lin2025full1.0, lin2025full1.5} and discerning emotional cues from speech \cite{wu2024emo,Wu_2024}. While these efforts address important aspects of dialogue, they do not systematically evaluate \textit{how well an SDM can detect and respond to emotions in a context-aware manner.}

To bridge this gap, we introduce EMO-Reasoning, a benchmark designed to assess emotional reasoning, the system’s ability to detect, interpret, and incorporate a user’s emotional state from tone and wording into its responses, in existing SDMs. Our framework integrates both categorical (e.g., happy, angry) \cite{ekman1992argument} and dimensional (e.g., arousal, valence) \cite{russell1977evidence} representations to capture a wide spectrum of emotional phenomena. In the categorical paradigm, we measure how closely an agent’s emotional expressions match those of the user at each dialogue turn. In the dimensional paradigm, we focus on the dynamics of emotion over time, employing metrics such as emotional contagion and balancing to examine whether an agent’s responses remain coherent across varying affective states. To enable detailed continuous emotion analysis, we develop a continuous speech emotion recognition (CSER) model for frame-level predictions, complementing a discrete emotion classifier for categorical labels.

Based on our experiments, surprisingly, we observe that existing SDMs struggle to deliver strong emotional reasoning, revealing opportunities to further refine these models. We summarize our contributions as follows:
\begin{itemize}[topsep=0pt, noitemsep, leftmargin=*]
\item We collect a multi-turn emotional dialogue dataset via an expressive text-to-speech model, covering diverse emotional contexts.
\item We propose novel metrics that consider categorical and dimensional perspectives for a holistic evaluation of emotional reasoning.
\item We present a CSER model that predicts fine-grained emotion trajectories in speech, enabling continuous analysis of affective states.
\item We provide a new benchmark for emotional reasoning capabilities in SDMs, highlighting important directions for improvement.
\end{itemize}

%% file: tex/method.tex
\section{EMO-Reasoning Framework}

\begin{figure*}[t]
  \centering
  \includegraphics[width=\textwidth]{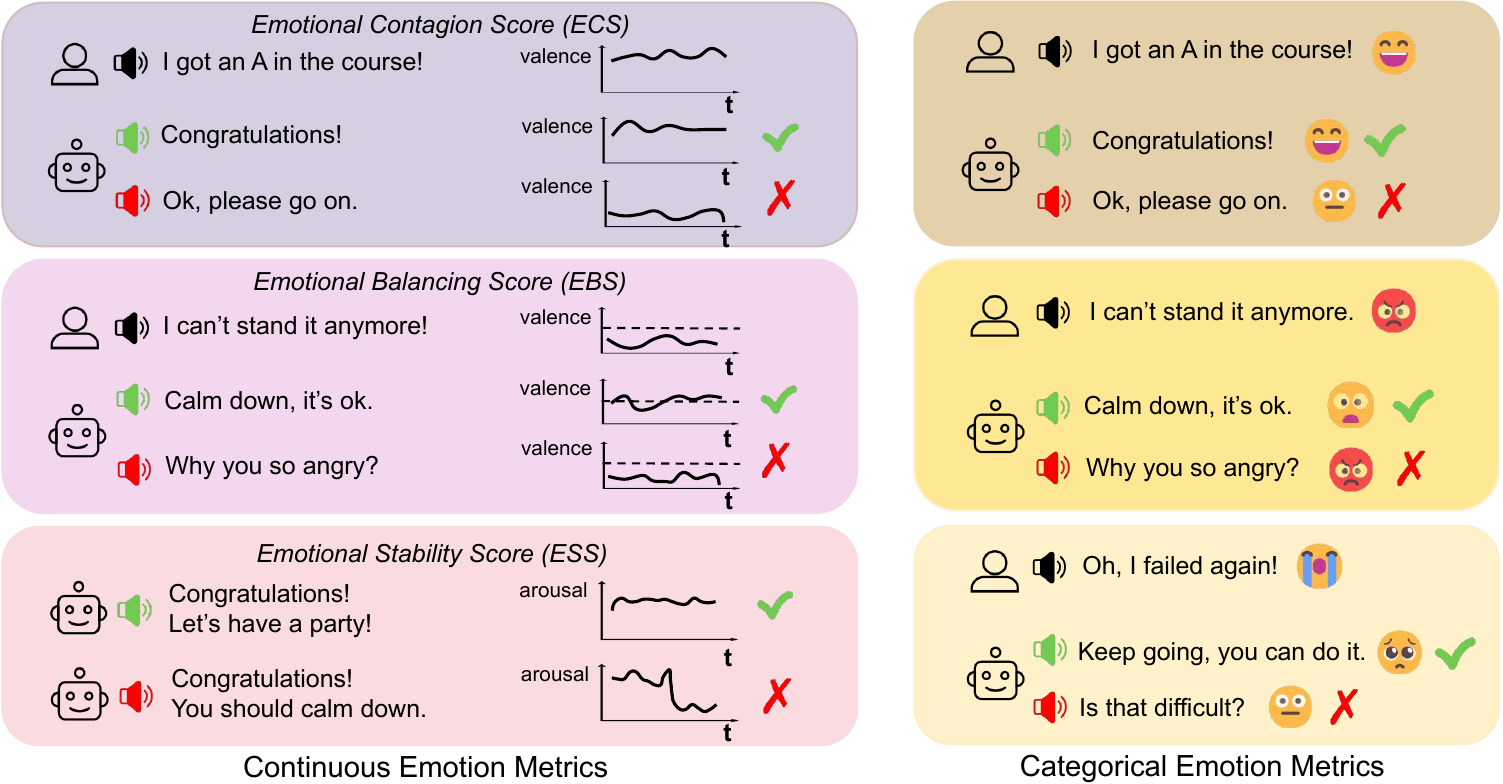}
  \caption{{\bf Overview of Emotion Reasoning Evaluation.} Our benchmark assesses how each spoken dialogue model responds to user queries using either continuous (left) or categorical (right) emotion metrics.}
  \label{fig:metrics}
\end{figure*}

We evaluate emotional reasoning in two ways: (1) continuous emotion metrics; and (2) categorical emotion metrics (Figure~\ref{fig:metrics}). All metrics are normalized to the range [0, 1] using dataset‑level bounds, where higher values imply better performance. This enables direct comparison and averaging without favoring models with wider raw ranges. Four high‑agreement categories (neutral, happy, angry, sad) keep discrete evaluation reliable and balanced. Nuance is supplied by continuous arousal/valence/dominance trajectories and human perceptual ratings, covering subtle gradations without low‑reliability classes.


\subsection{Evaluation Metrics for Continuous Emotions}
\label{ss:continuous_emotion}
Our continuous metrics rely on frame-level valence, arousal, and dominance (VAD) predictions. In each formula, $E$ represents each emotion value V/A/D, $\text{norm}(\cdot)$ indicates a [0,1] normalization, and $\text{DTW}(\cdot, \cdot)$ applies dynamic time warping.

\subsubsection{Single-turn Metrics}
We compute Emotional Contagion Score (ECS), Emotional Balancing Score (EBS), Emotional Stability Score (ESS), and Emotion Reasoning Score (ERS) for valence (V)/arousal (A)/dominance (D), separately, and compute the final value by averaging them.\\
\mypar{Emotional Contagion Score (ECS).}
Emotional contagion refers to how a speaker’s emotional state influences the listener's emotions. When users express happiness, it is expected that SDMs will also exhibit a happy response. Therefore, ECS measures whether the SDMs’ valence ($V_m$) and arousal ($A_m$) mirror the user's $(V_u, A_u)$:

\begin{align}
  \text{ECS} &= \text{norm}(-(\text{DTW}(V_m, V_u) + \text{DTW}(A_m, A_u))).
\end{align}

This metric is particularly useful for evaluating appropriate mirroring of positive emotions, such as happiness.


\mypar{Emotional Balancing Score (EBS).}
EBS evaluates how well the system moderates extreme emotions in a user, aiming to reduce emotional extremes (sadness, anger, etc.). 
EBS rewards de‑escalation only when user affect is extreme and negative; otherwise, it is inert. When an extreme emotion is detected, we measure how closely the system’s trajectory of the machine's emotion $E_m$ follows a desired balanced path $(E_u + \delta_E)$. $E_u$ here denotes the user's emotion. $\delta_E$ is the expected balance of V/A/D, which is the degree to which we hope the machine's emotion $E_m$ will be pulled back to a stable state compared to the user's extreme emotions. We sum the DTW distance across valence, arousal, and dominance, then normalize:
\begin{align}
\text{EBS} &= 
  \text{norm}(-\sum_{E} \text{DTW}(E_u + \delta_E, E_m) \cdot I_E) \quad
\end{align}
, where an indicator function \(I_E\) is used to check whether each emotional dimension is extreme. If no extreme emotion occurs, we set $\text{EBS}=0$. Frame/window V/A/D predictions aligned with DTW retain the temporal shape (gradual calming vs. abrupt spikes) that utterance averages discard. This exposes pacing and regulation quality, which is central to emotional reasoning. We use DTW with detailed predictions because emotion is a dynamic process. An utterance-level score would miss important shifts that occur within a conversation turn. By analyzing the entire emotional trajectory, DTW provides a more sensitive and accurate evaluation of emotional alignment compared to a single overall score.

In contrast to ECS, which rewards mirroring positive affect, EBS is specifically designed to assess the moderation of intense negative emotions, such as extreme anger, which is an important empathetic skill. The defined ECS measures the appropriate reflection of positive emotions, like happiness, within our four-class emotion recognition system. These metrics offer a more nuanced evaluation of contextual appropriateness.

\mypar{Emotional Stability Score (ESS).}
ESS evaluates the agent's ability to maintain consistent emotional control. We track $\Delta E(t) = E(t+1) - E(t)$ and penalize large jumps exceeding a threshold $\delta$, where $E(t)$ denotes the emotion value of V/A/D at time $t$. We sum these penalties across the dialogue and normalize:
\begin{align}
\text{ESS} &= \text{norm}(-\sum_{E} \sum_{t=0}^{T-1} |\Delta E(t)| \cdot (|\Delta E(t)| > \delta)) \quad .
\end{align}

\mypar{Emotion Reasoning Score (ERS).}
ERS is the average of ECS, EBS, and ESS. If no extreme user emotions occur, EBS is ignored, and ERS becomes the mean of ECS and ESS. The final ERS score is the average of ERS for all V/A/D.

\subsubsection{Cross-turn Metrics}
For multi-turn dialogues, we extend from the single-turn metrics:

\mypar{Cross-turn Emotional Contagion Score (CT-ECS).}
We average the ECS over all turns to capture long-term contagion as CT-ECS.

\mypar{Cross-turn Emotional Balancing Score (CT-EBS).}
We average the EBS only for turns with extreme user emotions as CT-EBS. If no extreme emotion arises, CT-EBS is 0.

\mypar{Cross-turn Emotional Stability Score (CT-ESS).}
The CT-ESS captures stability across turns by measuring the DTW distance between consecutive turns:
\begin{align}
\text{CT-ESS} &= \text{norm}(-\sum_{n=1}^{N-1} \sum_{E} \text{DTW}(E_n, E_{n+1})) \quad
\end{align}
, where \(E_n\) denotes the system’s  trajectory of the emotion value V/A/D at turn \(n\) in a multi-turn conversation.

\mypar{Cross-turn Emotion Reasoning Score (CT-ERS).}
Similarly, CT-ERS averages CT-ECS, CT-EBS, and CT-ESS. If no extreme emotion is detected, we ignore CT-EBS, and CT-ERS is the mean of CT-ECS and CT-ESS. The final CT-ERS score is the average of the CT-ERS for all V/A/D.

\subsection{Evaluation Metrics for Categorical Emotions}
We focus on four emotion categories: neutral, happy, angry, and sad. This choice ensures high label reliability and a balanced evaluation set. To capture a broader and more nuanced spectrum of affective states, this categorical analysis is complemented by our continuous emotion metrics (Arousal/Valence/Dominance) in Section~\ref{ss:continuous_emotion}.

Table~\ref{tab:emotion_reasoning} shows our categorical emotion reasoning scores (ERS) (1--10 scale normalized to [0,1]) rated by 20 human evaluators based on the rationality of user–SDM emotion label pairs in their understanding of real dialogues. Each cell $(i,j)$ is the average rationality score when the user’s emotion label is $i$ and the SDM response's emotion label is $j$. For example, the entry at row “Happy,” column “Sad” is 0.2, meaning that if the user’s emotion is happy and the SDM responds with sad, the categorical ERS is 0.2. The final score for a multi-turn dialogue is the average ERS across all turns.


\begin{table}[t]
 \scriptsize
  \caption{{\bf Categorical Emotion Reasoning Scores.} The categorical ERS scores are rated by 20 human evaluators. Rows denote the user's emotion labels, while columns indicate the corresponding emotion responses generated by SDMs.}
  \label{tab:emotion_reasoning}
  \centering
  \begin{tabular}{lccccc}
    \toprule
    \textbf{} & \textbf{Neutral} & \textbf{Happy} & \textbf{Angry} & \textbf{Sad} \\
    \midrule
    \textbf{Neutral} & 0.9 & 0.6 & 0.3 & 0.4 \\
    \textbf{Happy}   & 0.5 & 1.0 & 0.2 & 0.2 \\
    \textbf{Angry}   & 0.8 & 0.1 & 0.4 & 0.5 \\
    \textbf{Sad}     & 0.6 & 0.2 & 0.4 & 0.9 \\
    \bottomrule
  \end{tabular}
\end{table}

\subsection{Human Perceptual Evaluations}

We complemented and enhanced our automatic metrics by incorporating human ratings on a five-point scale (1 = worst, 5 = best). 
Then, the ratings were normalized to the range [0,1] and assessed across three dimensions: \textbf{Emotional Rationality (ER), Emotional Naturalness (EN), and Response Relevance (RR).} 
For each dialogue, evaluators judged how rational, natural, and relevant the emotional response of each SDM was, based on 20 randomly sampled dialogues.

For this evaluation, we recruited 20 human annotators (10 native and 10 non-native English speakers). 
Each annotator rated 20 randomly sampled dialogues for all seven SDMs as well as the ground-truth human responses from DailyTalk. Their task was to evaluate each response’s rationality, naturalness, and relevance.

\mypar{Emotional Rationality (ER).}
ER measures how appropriate the SDM’s emotional response is to the user’s speech (1 = very unreasonable, 5 = very reasonable). ER does not simply evaluate emotional matching but assesses the agent’s understanding and empathy. For example, if the user is very happy, the SDM may mirror that happiness; if the user is angry, the SDM should express empathy and comfort rather than reciprocating the anger.

\mypar{Emotional Naturalness (EN).}
EN evaluates how natural and smooth the SDM’s emotional expression is (1 = very unnatural, 5 = very natural), indicating the absence of any mechanical or forced tone. A high score indicates that the SDM's emotional responses flow smoothly.

\mypar{Response Relevance (RR).}
RR assesses how well the system’s utterance addresses the user’s emotional and contextual needs (1 = very irrelevant, 5 = very relevant). Responses that are more detailed, specific, and contextually rich will receive higher scores. For instance, when the user expresses sadness, a simple ``don’t be sad'' might be less relevant than a more empathetic response that acknowledges the user’s feelings and offers comfort.

\mypar{Emotion Reasoning Score (ERS).}
We average these three scores to obtain the final human-perceived Emotion Reasoning Score (ERS).



%% file: tex/experiment.tex
\section{Continuous Speech Emotion Recognition}
To better capture fine-grained emotional trajectories in speech for our proposed metrics, we train a Continuous Speech Emotion Recognition (\textbf{CSER}) model as detailed below:

\subsection{Training Dataset}
We use the MSP-Conversation dataset \cite{martinez2020msp} to train our CSER model. This dataset features multi-speaker podcast-style interactions, each annotated by at least six raters for valence, arousal, and dominance. We use version 1.1, which includes 224 conversations (59.6 hours). We follow the dataset’s recommended splits, using 148 conversations for training, 39 for development, and 57 for testing.

\subsection{CSER Model Configuration and Baseline}
\begin{figure}[t]
  \centering
  \includegraphics[width=0.9\linewidth]{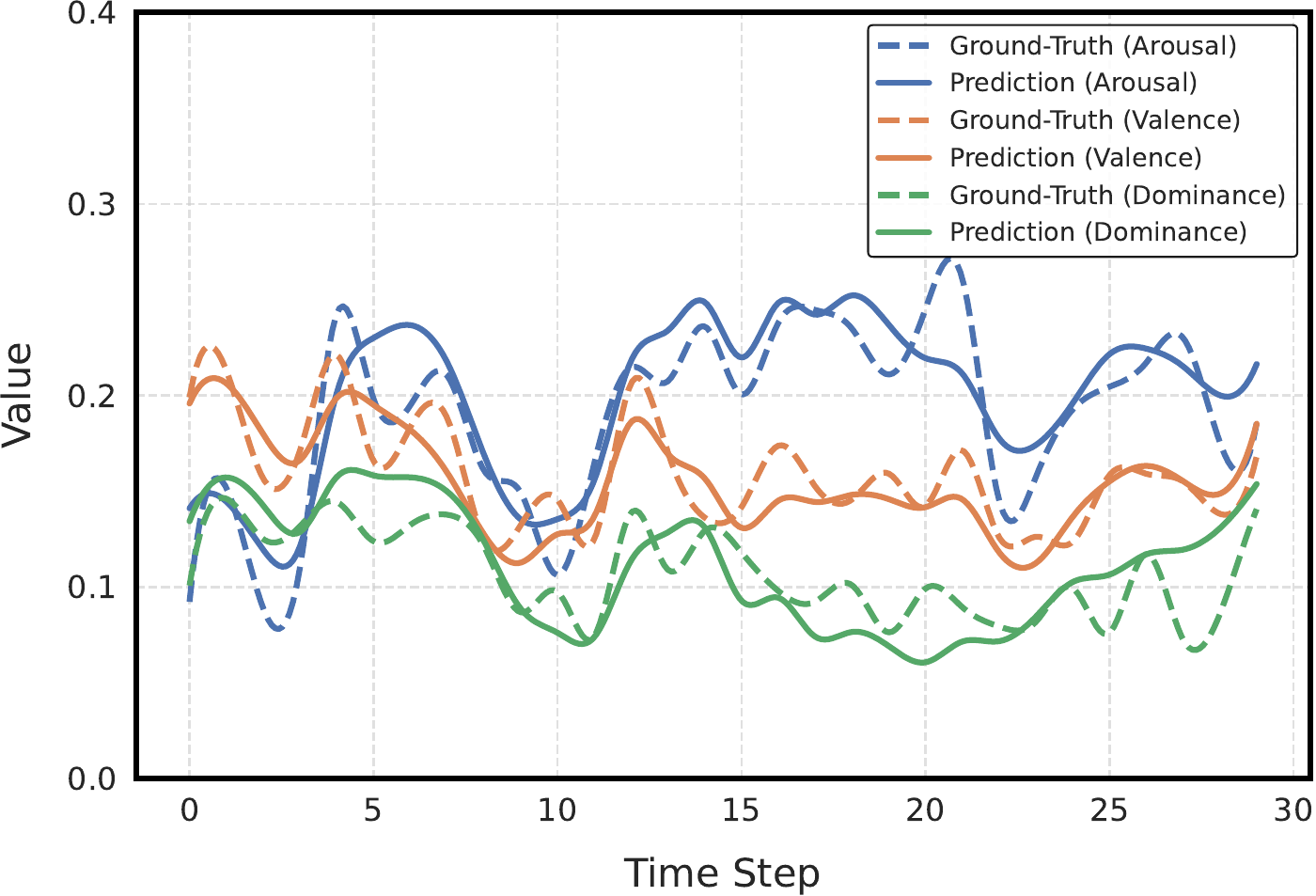}
  \caption{{\bf Continuous Emotion Predictions.} We show the example alignments between our prediction and the ground truth.}
  \label{fig:continuous_prediction}
\end{figure}

We segment the MSP-Conversation data into 20- to 50-second clips. Because the original annotations are provided at around 60Hz, we average them both across annotators and within each second, resulting in one label per second. We use WavLM-Large \cite{chen2022wavlm} to extract acoustic features, aggregated by second. Our model is a BiLSTM trained with the concordance correlation coefficient (CCC) loss \cite{martinez2024dynamic}. Table~\ref{tab:CSER_result} compares our BiLSTM with the baseline from \cite{martinez2024dynamic}. Fig.~\ref{fig:continuous_prediction} shows an example of our continuous predictions. We chose this architecture for its proven effectiveness in sequence modeling tasks. As shown in Table~\ref{tab:CSER_result}, our CSER model is competitive and outperforms the baseline from \cite{martinez2024dynamic} on valence and dominance, confirming its suitability as a feature extractor for our benchmark.

\section{Experimental Setup}

\begin{table}[t]
  \caption{Concordance correlation coefficient comparison between our method and BiLSTM baseline.}
  \label{tab:CSER_result}
  \scriptsize
  \centering
  \begin{tabular}{lccc}
    \toprule
    \textbf{Model} & Arousal & Valence & Dominance \\
    \midrule
    \textbf{BiLSTM} \cite{martinez2024dynamic} & 0.594 & 0.390 & 0.435 \\
    \textbf{Ours} & 0.554 & 0.449 & 0.539 \\
    \bottomrule
  \end{tabular}
\end{table}

\begin{table}[t]
\fontsize{7}{9}\selectfont
  \caption{Statistical distribution of prompts and their corresponding speech data.}
  \label{tab:prompt_stats}
  \centering
  \begin{tabular}{lcccccc}
    \toprule
    \textbf{} & \textbf{Neutral} & \textbf{Happy} & \textbf{Angry} & \textbf{Sad} &  \textbf{NLD} \\
    \midrule
    \textbf{\# Prompts} & 500 & 500 & 500 & 500 & 1251\\
    \textbf{Duration (h)}   & 1.26 & 1.17 & 1.15 & 1.34 & 2.45\\
    \bottomrule
  \end{tabular}
\end{table}

\subsection{Dataset}
\subsubsection{Synthetic Data}
Existing emotional speech datasets often lack rich contextual cues and display limited emotional variety. To address this, we design a text-to-speech pipeline.
We prompt a large language model (GPT-4) to produce dialogue text and emotional descriptors. These instructions range from basic labels (e.g., happy, angry) to more expressive natural language descriptions (NLD) such as ``Anger and despair, loud and sharp delivery, as if questioning the meaning of everything with a furious edge''. We also vary the number of dialogue turns for single- and multi-turn conversations, ensuring realistic emotional transitions (e.g., anger → calm → gratitude) and compatibility with multiple potential responses from the agent.

We employ CosyVoice \cite{du2024cosyvoice}, a TTS framework that supports fine-grained emotional control. Using the generated text and descriptors, we create high-quality emotional speech. Table~\ref{tab:prompt_stats} summarizes the prompt distribution and total audio duration.

\subsubsection{Real Data}
For real human speech evaluation, we use DailyTalk \cite{lee2023dailytalk}, a curated set of 2,541 dialogues (21.7 hours) adapted from the open-domain DailyDialog dataset \cite{li2017dailydialog}. DailyTalk offers high-quality recordings suitable for testing emotional responses under natural speaking conditions. We compare the emotional reasoning performance on DailyTalk against SDM's performance on dialogues generated by GPT-4 and CosyVoice (2566 single-turn and 311 multi-turn).


\subsection{Metric Hyperparameters}
We set hyperparameters based on the statistical properties of MSP-Conversation. For EBS, we define extremes using the 80th percentile of arousal (0.345) and the 20th percentile of valence (-0.07) and dominance (0.210). We also set \(\delta_A = -0.105, \delta_V = 0.211,  \delta_D = 0.098\) based on the 50th percentile. For ESS, we apply a threshold $\delta=0.04$ based on the 80th percentile of consecutive label changes. All component scores are normalized to a range of [0, 1] using min-max normalization, which employs empirically determined feasible bounds. All metric thresholds (extreme affect, stability jump) are fixed from dataset percentiles (not tuned per model), grounding them in empirical distributions. Rankings remain unchanged
under $\pm 5$ percentile shifts, and absolute score changes
are small $(< 0.02)$. 




\begin{table*}[t]
  \caption{{\bf Emotion reasoning performance in single-turn dialogues on the DailyTalk.} We evaluate proposed continuous, categorical, and perceptual metrics on several methods.}
  \label{tab:model_performance_single_turn}
  \scriptsize
  \centering
  \begin{tabular}{lcccccccccccccc}
    \toprule
    \multirow{2}{*}{\bf Model} & \multicolumn{4}{c}{\textbf{Continuous Based}} & \multicolumn{5}{c}{\textbf{Categorical Based}} & \multicolumn{4}{c}{\textbf{Perceptual Based}} \\
    \cmidrule(lr){2-5} \cmidrule(lr){6-10} \cmidrule(lr){11-14}
    & ECS & EBS & ESS & ERS & Neutral & Happy & Angry & Sad & Avg & ER & EN & RR & ERS \\
    \midrule
    Real \cite{lee2023dailytalk}  & 0.851 & 0.834 & 0.967 & 0.884 & 0.863 & 0.984 & 0.701 & 0.828 & 0.844 & 0.910 & 0.923 & 0.953 & 0.928 \\
    \midrule
    LLaMA-Omni \cite{fang2024llama}   & 0.655  & 0.614  & 0.669  & 0.646 & 0.601 & \textbf{0.968}  & 0.492  & 0.497 & 0.640 & 0.557 & 0.525 & 0.671 & 0.584 \\
    mini-omni \cite{xie2024mini} & 0.704 & 0.518 & 0.733 & 0.652 & 0.610 & 0.931 & 0.512 & 0.594 & 0.662 & 0.579 & 0.547 & 0.551 & 0.560 \\
    mini-omni2 \cite{xie2024miniomni2opensourcegpt4ovision} & 0.737 & 0.626 & 0.762 & 0.708 & 0.649 & 0.862 & \textbf{0.690} & 0.589 & 0.697  & 0.615 & 0.635 & 0.629 & 0.626 \\
    Freeze-Omni \cite{wang2024freeze} & \textbf{0.803} & 0.674 & 0.844 & 0.774 & 0.651 & 0.940 & \textbf{0.690} & \textbf{0.669} & \textbf{0.738} & \textbf{0.764} & \textbf{0.761} & 0.832 & \textbf{0.785} \\
    GLM-4-Voice \cite{zeng2024glm} & 0.795 & \textbf{0.713} & \textbf{0.902} & \textbf{0.803} & \textbf{0.779} & 0.849 & 0.682 & 0.596 & 0.726 & 0.756 & 0.634 & \textbf{0.879} & 0.756 \\
    Moshi \cite{defossez2024moshi} & 0.637 & 0.512 & 0.665 & 0.604 & 0.579 & 0.671 & 0.343 & 0.390 & 0.496 & 0.283 & 0.277 & 0.293 & 0.284 \\
    dGSLM \cite{nguyen2023generative} & 0.639 & 0.478 & 0.564 & 0.560 & 0.543 & 0.637 & 0.417 & 0.312 & 0.477 & 0.331 & 0.298 & 0.258 & 0.296 \\
    \bottomrule
  \end{tabular}
\end{table*}

\begin{table}[t]
  \caption{{\bf Emotion reasoning performance in multi-turn dialogues on the DailyTalk.} We measure metrics in multi-turn dialogues, where {\em Cate.} refers to categorical.}
  \label{tab:model_performance_multi_turn}
  \scriptsize
  \setlength{\tabcolsep}{0.52mm}{
  \centering
  \begin{tabular}{lcccccccccccccc}
    \toprule
    \multirow{2}{*}{\bf Model} & \multicolumn{4}{c}{\textbf{Continuous Based}} & \multicolumn{1}{c}{\textbf{Cate. Based}} & \multicolumn{4}{c}{\textbf{Perceptual Based}} \\
    \cmidrule(lr){2-5} \cmidrule(lr){6-6} \cmidrule(lr){7-10}
    & ECS & EBS & ESS & ERS & Avg & ER & EN & RR & ERS \\
    \midrule
    Real \cite{lee2023dailytalk} & 0.844 & 0.838 & 0.894 & 0.859 & 0.844  & 0.899 & 0.927 & 0.949 & 0.925 \\
    \midrule
    Moshi \cite{defossez2024moshi}   & \textbf{0.620}  & \textbf{0.523}  & \textbf{0.630} & \textbf{0.591}  & \textbf{0.470}  & 0.267 & 0.259 & \textbf{0.288} & 0.271 \\
    dGSLM \cite{nguyen2023generative} & 0.618 & 0.455 & 0.583 & 0.552 & 0.461 & \textbf{0.323} & \textbf{0.310} & 0.239 & \textbf{0.291} \\
    \bottomrule
  \end{tabular}}
\end{table}

\subsection{Agent Response Collection}
Using generated emotional prompts, we interact with seven SDMs: LLaMA-Omni \cite{fang2024llama}, mini-omni \cite{xie2024mini}, mini-omni2 \cite{xie2024miniomni2opensourcegpt4ovision}, Freeze-Omni \cite{wang2024freeze}, GLM-4-Voice \cite{zeng2024glm}, Moshi \cite{defossez2024moshi}, and dGSLM \cite{nguyen2023generative}. Moshi and dGSLM are further tested with multi-turn interactions since they support multi-turn dialogue generation.

\begin{itemize}[topsep=0pt, noitemsep, leftmargin=*]
  \item \textbf{LLaMA-Omni \cite{fang2024llama}}: Integrates a speech encoder, adapter, LLM, and streaming decoder to produce text + speech directly from raw input with minimal latency.

  \item \textbf{Mini-Omni \cite{xie2024mini}}: Uses a MusicGen‐style decoder to jointly generate text and speech tokens via padded alignment and batch‐parallel inference.

  \item \textbf{Mini-Omni2 \cite{xie2024miniomni2opensourcegpt4ovision}}: Builds on Mini-Omni by adding a cross‐modal encoder that fuses audio embeddings with dialogue history, enabling duplex (simultaneous receive/send) capability.

  \item \textbf{Freeze-Omni \cite{wang2024freeze}}: Freezes a pretrained ASR encoder and TTS decoder, then trains a lightweight bridge to map speech embeddings into the decoder, enabling speech‐to‐speech dialogue using only text–speech paired data.

  \item \textbf{GLM-4-Voice \cite{zeng2024glm}}: Built on GLM-4, this two‐stage system first generates text tokens, then produces speech codec tokens; an alignment predictor reduces idle time, though overall latency remains higher than streaming approaches.

  \item \textbf{Moshi \cite{defossez2024moshi}}: Employs a global‐local Transformer and multi‐layer neural codec: it predicts time-aligned text tokens as a prefix, then generates residual codec codes in parallel streams for user and system, removing explicit turn markers and jointly modeling semantics and acoustics.

  \item \textbf{dGSLM \cite{nguyen2023generative}}: An early end‐to‐end system trained on paired audio that uses self‐ and cross‐attention to map input waveforms directly to outputs, achieving natural interactivity without an LLM component.

\end{itemize}

These models span LLM‐centric streaming (LLaMA-Omni, GLM-4-Voice), codec‐based autoregressive (Mini-Omni, Mini-Omni2), and purely speech‐driven (Freeze-Omni, dGSLM, Moshi) approaches. Their performance on emotional prompts highlights how each handles affect in single‐turn (and multi‐turn, where applicable) dialogue.


\subsection{Evaluation}

We then apply our continuous and categorical metrics on the DailyTalk dataset. For continuous measurements, we use our CSER model to track valence, arousal, and dominance for both user and agent speech. For categorical labels, we use SenseVoice \cite{an2024funaudiollm}, a state-of-the-art emotion classification model. Human annotators also rate each dialogue’s emotional performance to provide a quasi-ground-truth perceptual score.

\section{Results}

\begin{figure}[t]
  \centering  \includegraphics[width=0.9\linewidth]{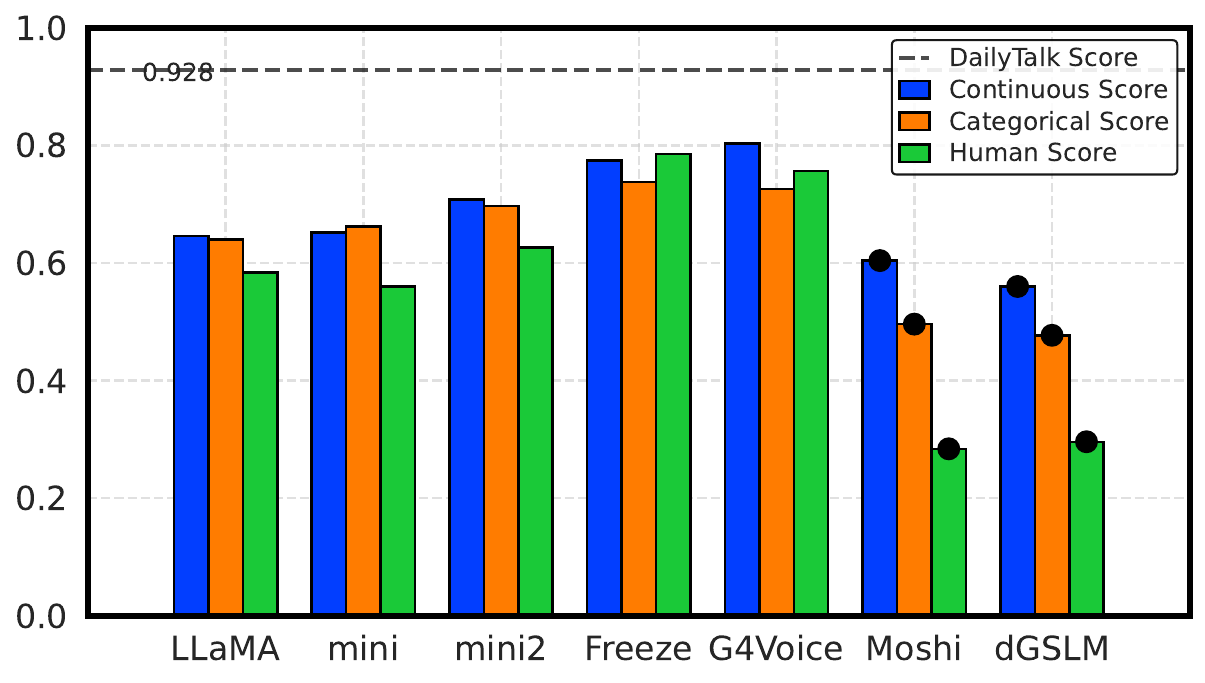}
  \caption{{\bf Metric Comparison.} The horizontal dotted line indicates the human ERS on DailyTalk. Black markers represent multi-turn scores for Moshi and dGSLM under various evaluation metrics.}
  \label{fig:score_compare}
\end{figure}

\begin{figure}[t]
  \centering
  \includegraphics[width=0.8\linewidth]{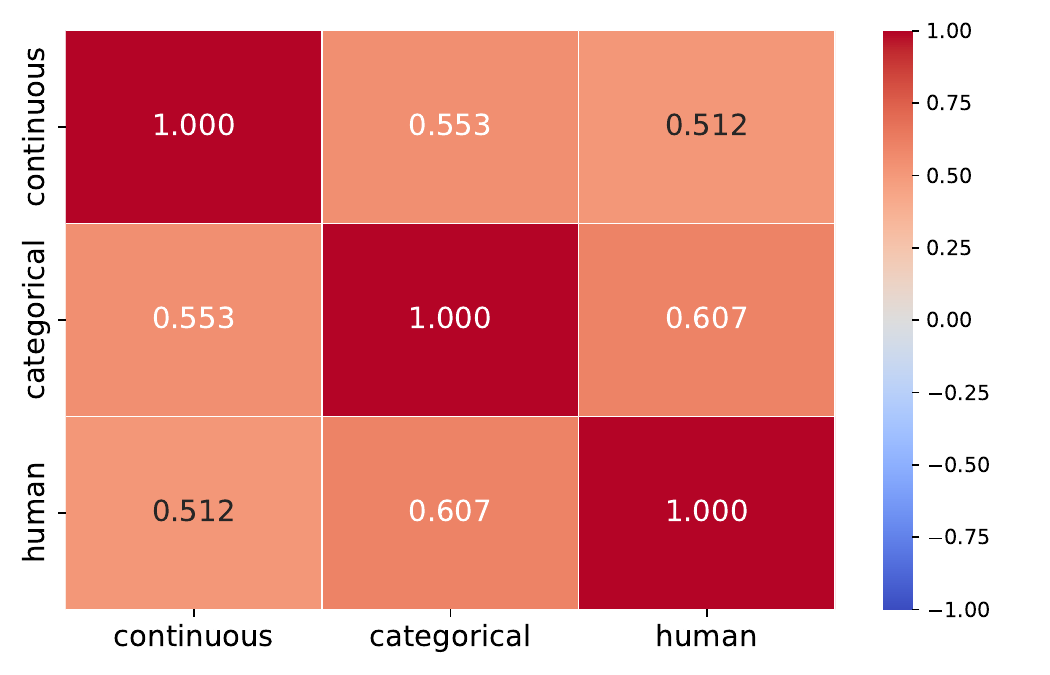}
  \caption{{\bf Metric correlation analysis.} We assess the correlations among continuous, categorical, and perceptual-based metrics.}
  \label{fig:correlation}
\end{figure}

\mypar{Automatic vs. Perceptual Metrics.}
To assess how well our automated measures reflect actual human judgments, we compare continuous and categorical scores against \textbf{human ERS} in Figures \ref{fig:score_compare} and ~\ref{fig:correlation}. In Figure 3, each bar represents a model’s performance under continuous score, categorical score, and human score, while the horizontal dotted line denotes the human ERS on DailyTalk. Notably, the continuous and categorical scores in single-turn responses from LLaMA-Omni, mini-Omni, mini-Omni2, Freeze-Omni, and GLM-4-Voice consistently approach the human score, demonstrating that our automated scores capture a substantial portion of what humans perceive as natural, emotionally coherent responses. However, when examining single-turn and multi-turn outputs from Moshi and dGSLM, a clear discrepancy emerges. Human evaluators assign these dialogues a relatively low ERS, often falling well below the dotted human baseline, yet our continuous and categorical metrics overestimate their emotional coherence, placing them near or even above the line. This mismatch indicates that, despite strong overall correlations, our proposed measures can fail to capture certain aspects of emotional nuance in extended interactions. Both Figure \ref{fig:score_compare} and ~\ref{fig:correlation} show that continuous and categorical metrics correlate well with human ratings (all correlations exceed 0.5). This suggests that our proposed metrics align closely with perceptual assessments of emotional quality.

\mypar{Real vs. Generated Dialogue Performance.}
In single-turn (Table~\ref{tab:model_performance_single_turn}) and multi-turn (Table~\ref{tab:model_performance_multi_turn}) settings, real human dialogues from DailyTalk outperform all SDMs across continuous-based, categorical-based, and perceptual-based metrics. These results highlight that, while modern, advanced models capture some aspects of emotional reasoning, there remains a notable gap between generated responses and the nuanced coherence found in genuine human dialogues.

\mypar{Model Trade-offs in Single-Turn Dialogues.}
Freeze-Omni and GLM-4-Voice show relatively stronger single-turn performance than other SDMs, although each excels in different categories. Freeze-Omni achieves higher categorical-based scores, whereas GLM-4-Voice attains stronger continuous-based results. This indicates that optimizing for one aspect of emotional reasoning can introduce trade-offs in others.

\mypar{Challenges in Multi-Turn Emotional Consistency.}
Transitioning from single-turn exchanges to multi-turn dialogues reveals a pronounced decline in performance for both Moshi and dGSLM, as shown in Table~\ref{tab:model_performance_multi_turn}. In particular, their continuous-based scores (ECS, EBS, ESS, ERS) drop noticeably compared to single-turn settings, and their categorical averages similarly fall below expectations. Moreover, perceptual-based evaluations (ER, EN, RR, ERS) also register lower values than those achieved by real human conversations. Together, these reductions across continuous, categorical, and perceptual metrics underscore the inherent difficulty that current models face when attempting to preserve coherent emotional trajectories over several back-and-forth turns. Such performance drops can be attributed to the compounding complexity of maintaining context: each successive turn requires the model not only to generate a contextually appropriate response but also to preserve an emotional thread that aligns with prior utterances. In multi-turn scenarios, an early deviation in emotion or tone can propagate through subsequent responses, leading to inconsistencies that automated metrics detect—and that human raters penalize even more harshly. As a result, Moshi and dGSLM exhibit lower overall emotional alignment compared to real dialogues, highlighting the challenge of sustaining nuanced affective states over extended interactions. These findings validate why EMO-Reasoning places a strong emphasis on evaluating longer conversational segments rather than isolated single exchanges. By focusing on extended interactions, EMO-Reasoning exposes weaknesses in current SDM architectures and evaluation methods, bringing attention to the gap between model-generated and human-level emotional fluency. Consequently, these results validate the importance of EMO-Reasoning’s focus on extended interactions and emphasize the need for further research on emotional consistency in spoken dialogue systems.

%% file: tex/conclusion.tex
\section{Conclusion and Future Works}
\mypar{Conclusion.}
In this paper, we introduced \emph{EMO-Reasoning}, a new benchmark for evaluating emotional reasoning in spoken dialogue systems. Our framework combines continuous (valence, arousal, dominance) and categorical (happy, angry, sad, neutral, natural language descriptions) metrics to capture a broad range of emotional phenomena. We further developed a Continuous Speech Emotion Recognition (CSER) model, enabling fine-grained temporal analyses of emotional expression. Results on both single-turn and multi-turn scenarios demonstrate that while some advanced systems achieve moderate performance, they fall short of human-level emotional coherence, especially in extended interactions. We will release the full dataset and benchmark upon acceptance, and we hope the community will adopt and build upon EMO-Reasoning to accelerate research on emotionally aware spoken dialogue models.

\mypar{Limitation and Future Works.}
These findings underscore the importance of better modeling affective dynamics and designing mechanisms that maintain emotional consistency across turns. 
We will provide further details on inter-rater agreement and related analyses.
Future work could explore several directions. 
First, methods that unify discrete and continuous representations more seamlessly may improve alignment with human perception. 
Second, incorporating context-aware modulation (i.e., personal background or cultural norms) could help systems respond more appropriately to varying emotional states. 
Third, more robust training strategies that address emotional shifts over long dialogues may elevate multi-turn performance. 
By advancing each of these areas, we believe that next-generation spoken dialogue models can offer more empathetic and context-aware user experiences, ultimately narrowing the gap between current technology and natural human conversation. 
Lastly, in spoken health domains, modeling continuous rather than categorical emotion may provide alternative or salient cues for identifying dysfluencies~\cite{lian2023unconstrained-udm, lian-anumanchipalli-2024-towards-hudm, ssdm, lian2024ssdm2.0, zhou2024yolostutterendtoendregionwisespeech, zhang2025analysisevaluationsyntheticdata, guo2025dysfluentwfstframeworkzeroshot, ye2025seamlessalignment-neurallcs, zhou2024timetokensbenchmarkingendtoend, zhou2024stutter}, thereby enhancing the analysis and treatment of speech and language disorders. 
Also, we will investigate clinically grounded emotional representations to establish a universal biomarker and other languages.

\section{Acknowledgements}
We thank Luz Martinez-Lucas and Jingyao Wu for their discussions on training continuous emotion recognition models.